\documentclass[10pt,twocolumn,letterpaper]{article}

\usepackage{cvpr}
\usepackage{times}
\usepackage{epsfig}
\usepackage{graphicx}
\usepackage{amsmath, subfigure}
\usepackage{amssymb}
\usepackage{kotex}
\usepackage{makecell}
\usepackage{adjustbox}  
\usepackage{multirow}
\usepackage{xpunctuate} 
\usepackage{foreign}
\usepackage{pbox}
\usepackage{tabularx,booktabs,calc}
\usepackage{hhline}
\usepackage{enumitem}
\usepackage{stackengine}
\newcommand\xrowht[2][0]{\addstackgap[.5\dimexpr#2\relax]{\vphantom{#1}}}

\usepackage[breaklinks=true,bookmarks=false]{hyperref}

\cvprfinalcopy 

\def\cvprPaperID{****} 

\begin{document}

\title{ScarfNet: Multi-scale Features with Deeply Fused and Redistributed Semantics for Enhanced Object Detection}

\author{Jin Hyeok Yoo, Dongsuk Kum$*$, and Jun Won Choi\\
Hanyang University\\
Korea Advanced Institute of Science and Technology$*$
}

\maketitle

\begin{abstract}
Convolutional neural networks (CNNs) have led us to achieve significant progress in object detection research. To detect objects of various sizes, object detectors often exploit the hierarchy of the multiscale feature maps called {\it feature pyramids}, which are readily obtained by the CNN architecture. However, the performance of these object detectors is limited because the bottom-level feature maps, which experience fewer convolutional layers, lack the semantic information needed to capture the characteristics of the small objects. To address such problems, various methods have been proposed to increase the depth for the bottom-level features used for object detection. While most approaches are based on the generation of additional features through the top-down pathway with lateral connections, our approach directly fuses multi-scale feature maps using bidirectional long short-term memory (biLSTM) in an effort to leverage the gating functions and parameter-sharing in generating deeply fused semantics. The resulting semantic information is redistributed to the individual pyramidal feature at each scale through the channel-wise attention model. We integrate our semantic combining and attentive redistribution feature network (ScarfNet) with the baseline object detectors, i.e., Faster R-CNN, single-shot multibox detector (SSD), and RetinaNet. Experimental results show that our method offers a significant performance gain over the baseline detectors and outperforms the competing multiscale fusion methods in the PASCAL VOC and COCO detection benchmarks.
\end{abstract}

\section{Introduction}
Object detection refers to the task of deciding whether there are any instances of objects in the image and returns the estimate of the location and the category of the objects \cite{general}, \cite{survey}.
Historically, object detection has been one of the most challenging computer vision problems. Recently, deep learning has led unprecedented advances in object detection techniques \cite{survey}. Convolutional neural networks (CNNs) can produce a hierarchy of abstract feature maps through a cascade of convolution operations followed by a nonlinear function. Using CNN as a backbone network, object detectors can effectively infer the location of the bounding box and the category of the instances based on the abstract feature maps. 
Various object detection network structures have been proposed to date. The CNN-based object detectors are roughly categorized into two groups: two-stage detectors and single-stage detectors. The two-stage detectors detect the objects using two separate subnetworks; 1) the region proposal network for finding the bounding boxes containing the object, and 2) the object classifier network for identifying the class of the objects and refining the bounding boxes. The well-known two-stage detectors include R-CNN \cite{rcnn}, Fast R-CNN \cite{fastrcnn}, Faster R-CNN \cite{fasterrcnn}, and Mask R-CNN \cite{maskrcnn}. Single-stage detectors directly estimate the bounding boxes and the object classes from the feature maps in one shot and include single-shot multibox detector (SSD) \cite{ssd},  YOLO \cite{yolov1}, YOLOv2 \cite{yolov2}, and RetinaNet \cite{retinanet}. 

\begin{figure*}[!t]
	\centering
	\begin{subfigure}[]{\includegraphics[width=0.22\textwidth]{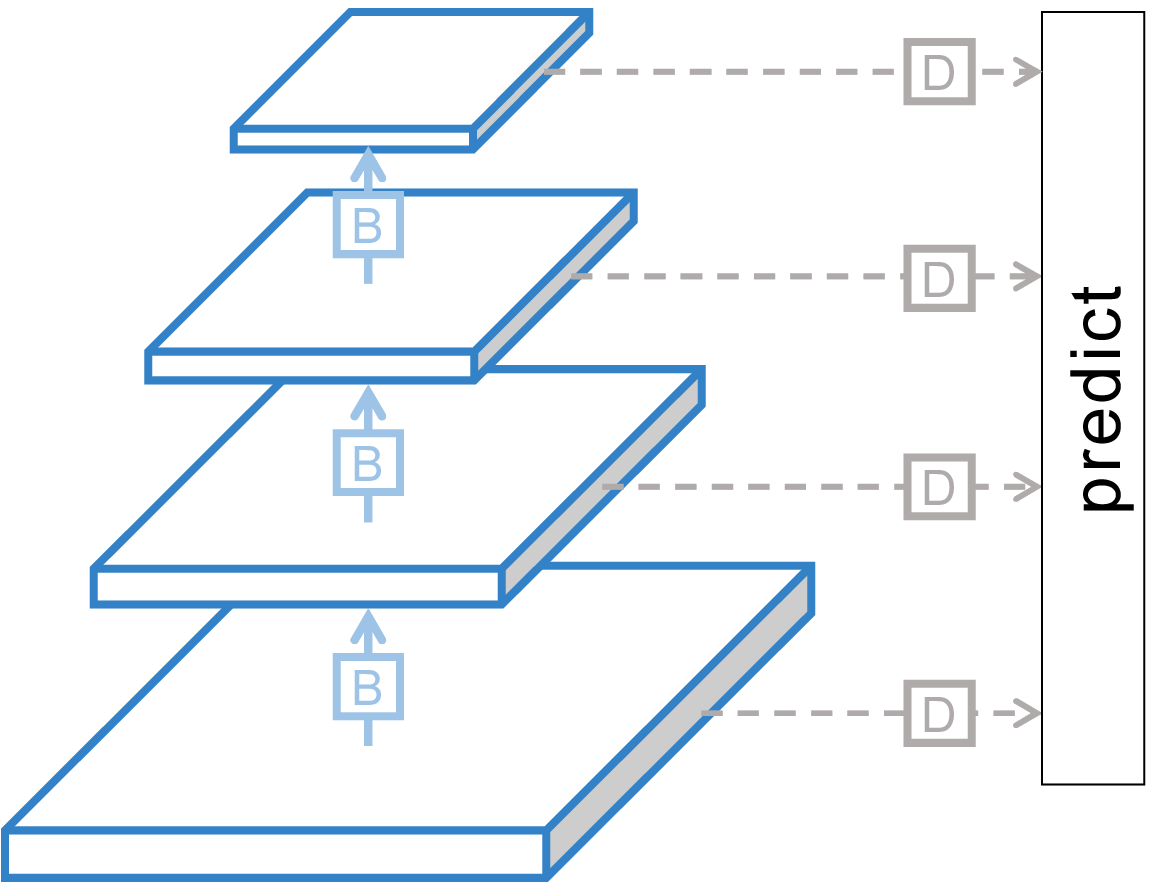}}
    \end{subfigure}
    \hspace{0.13cm}
	\begin{subfigure}[]{\includegraphics[width=0.35\textwidth]{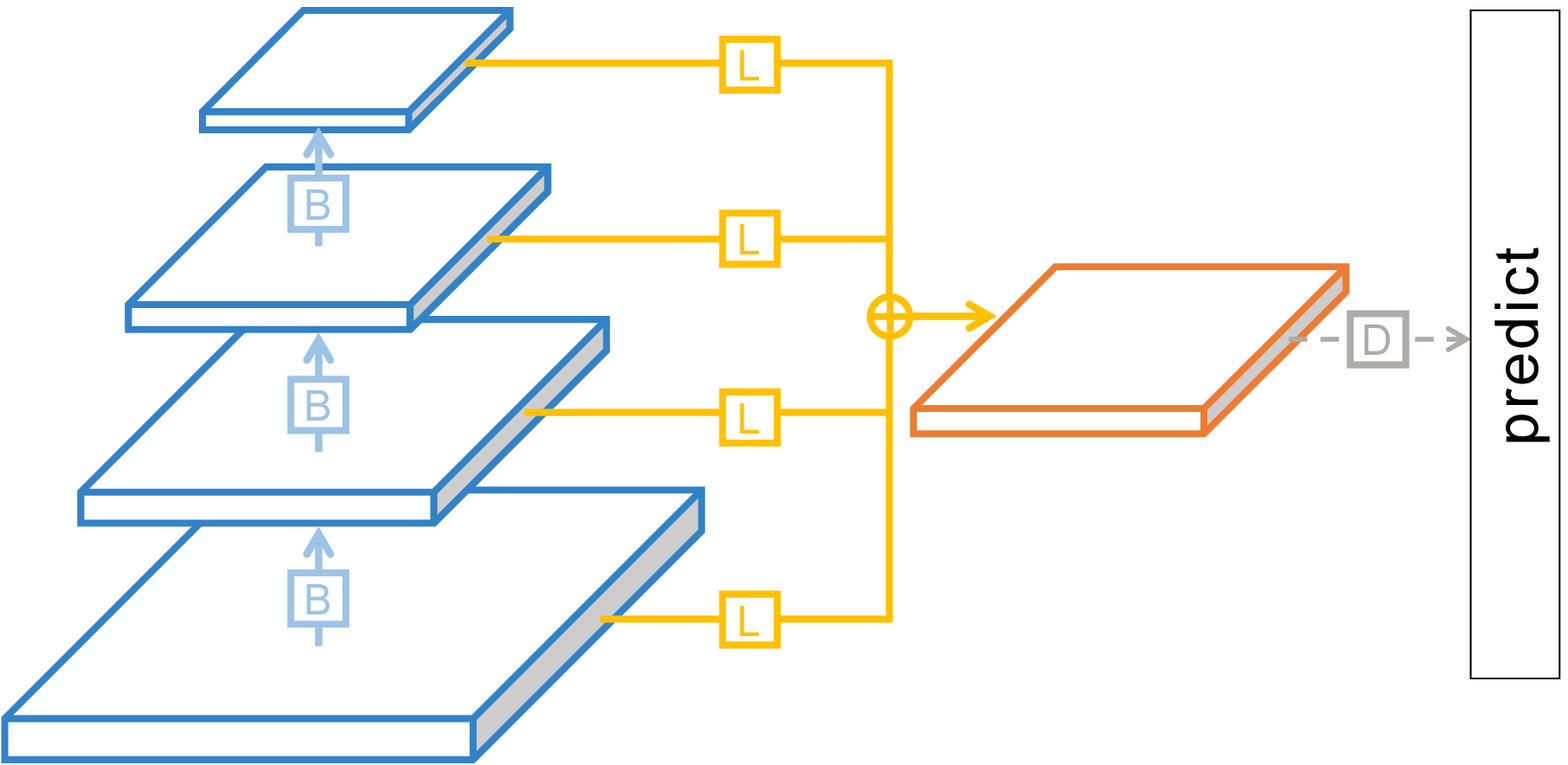}}
	\end{subfigure}
	\hspace{0.13cm}
	\begin{subfigure}[]{\includegraphics[width=0.38\textwidth]{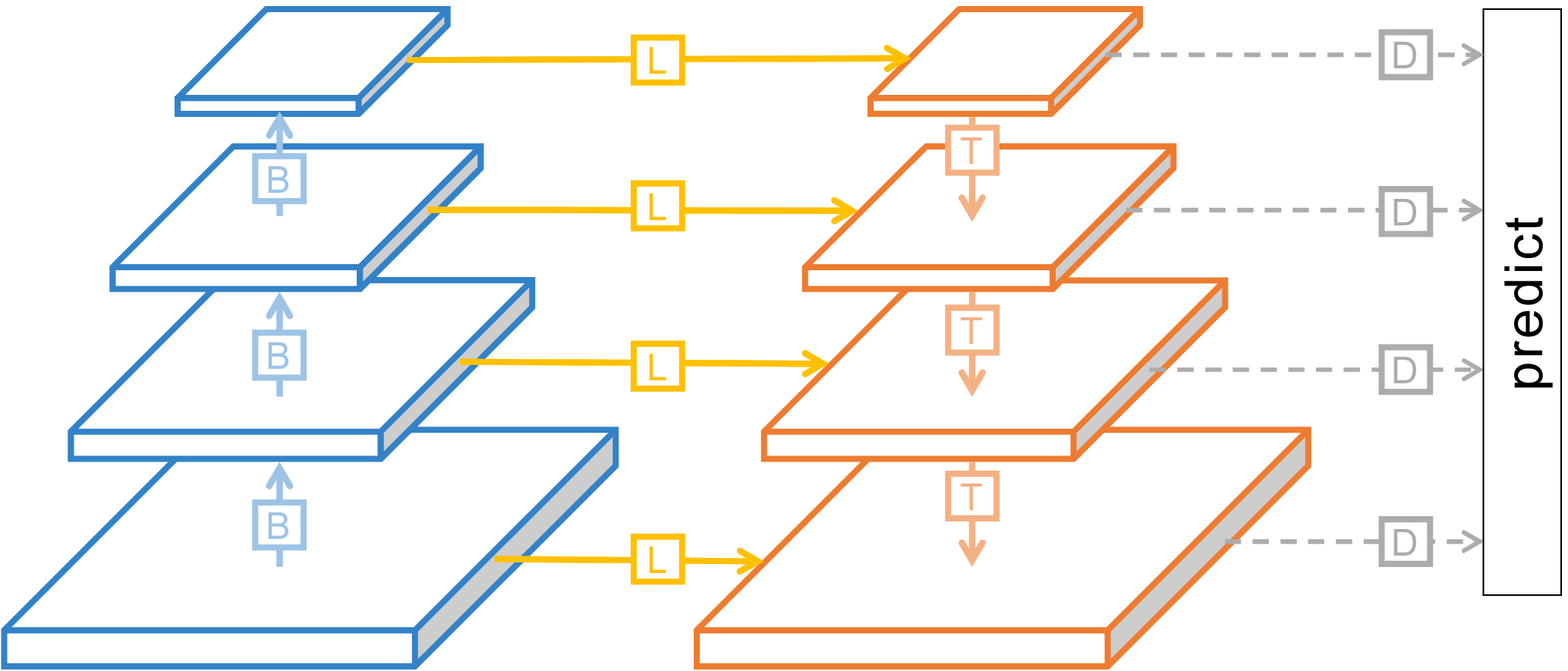}}
	\end{subfigure}\\ \vspace{0.3cm}\hspace{3cm}
	\begin{subfigure}[]{\includegraphics[width=0.56\textwidth]{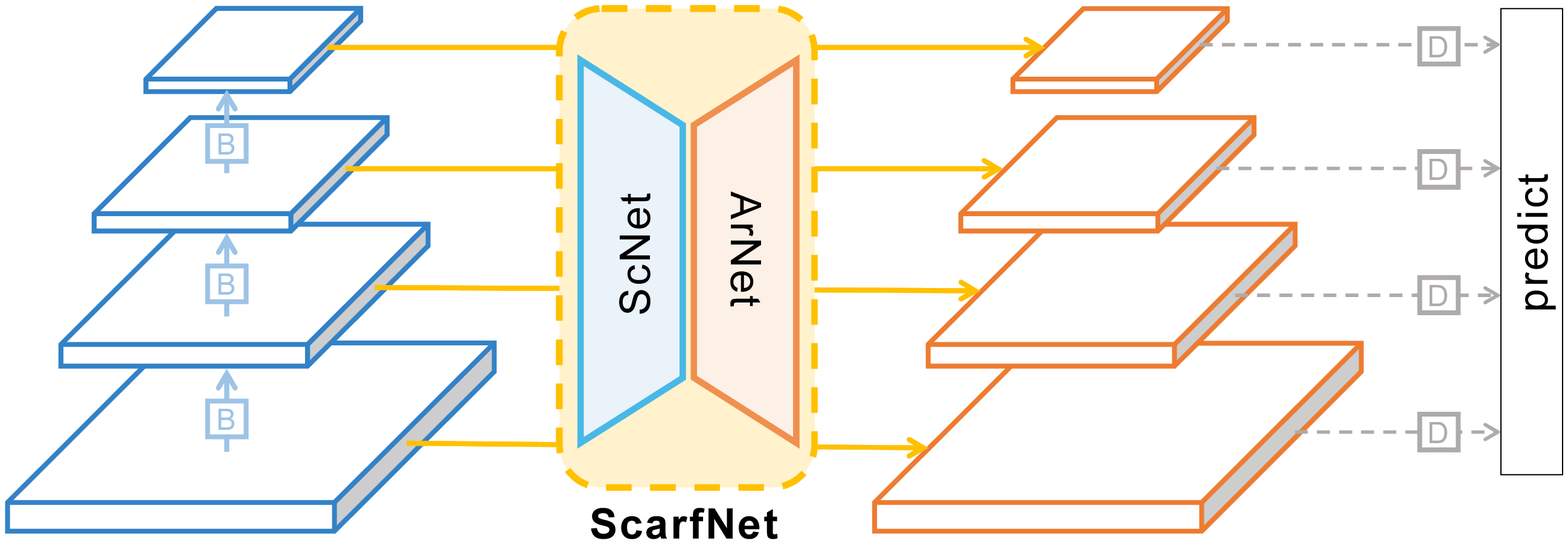}}
	\end{subfigure}
    \hspace{0.5cm}
	\begin{subfigure}[]{\includegraphics[width=0.2\textwidth]{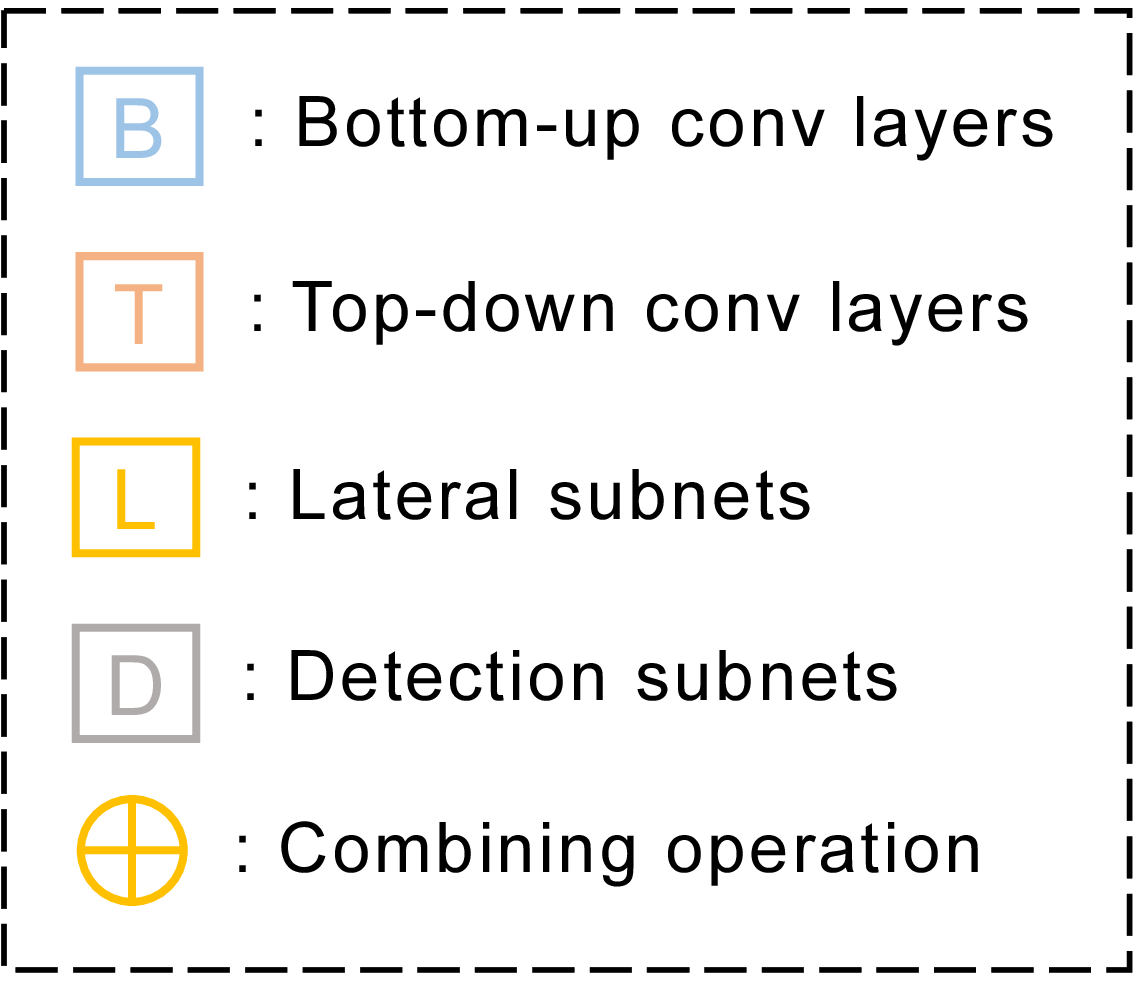}}
	\end{subfigure}
	\vspace{0.3cm}
	\caption {{\bf Structure of several feature pyramid methods:} In (a), feature pyramid obtained from convolutional layers is used in baseline detectors (e.g., SSD \cite{ssd}). In (b), multiscale features are fused and converted into the single semantic feature map with the highest resolution. (c) shows the structure generating additional features in unidirectional way through the top-down structure with lateral connections. (d) shows the structure of the proposed ScarfNet, where the multiscale features are fused in a bidirectional fashion and the learned semantics are propagated back to each scale.\vspace{-0.17cm}}
	\label{fig:comparison}
\end{figure*}

Recent advances in object detection are achieved by the CNN's capability to produce the abstract features containing strong semantic cues. The deeper the convolutional layers are, the higher the level of abstraction is for the resulting feature maps. As a result, the features produced at the end of the CNN pipeline (called {\it top-level features}) contain rich semantics but lack spatial resolution, whereas the features placed at the input layers (called {\it bottom-level features}) lack semantic information but have detailed spatial information. The hierarchy of such multiscale features constitutes so-called {\it feature pyramids}, which are used to detect the objects of different scales in many object detectors (e.g., SSD \cite{ssd}, MS-CNN \cite{cai2016unified}, and RetinaNet \cite{retinanet}). The structure designed to use such feature pyramids for object detection is described in Fig. \ref{fig:comparison} (a).
Note that the attributes of the large objects tend to be captured on the top-level features of small size while those of the small objects are well represented by the shallow bottom-level features of large size.

One limitation of the feature pyramid method is the disparity of the semantic information between the multiscale feature maps used for object detection. The bottom-level features are not deep enough to exhibit high-level semantics underlying in the objects and their surroundings. This results in the accuracy loss in detecting the small objects. In order to address this problem, several approaches, which have attempted to reduce the semantic gap between the pyramidal features in different scales, have been proposed. One notable direction is to provide the contextual information to the bottom-level features by generating the highly semantic features in the top-down pathway with latent connections. As illustrated in Fig. \ref{fig:comparison} (c), based on the top-level pyramidal feature obtained from the bottom-up network, the additional features are subsequently generated with increased depth and resolution. In order to avoid losing the spatial information, lateral connections are used to take the bottom-level features and combine them with the high-level semantic features. Various object detectors including DSSD \cite{dssd}, FPN \cite{fpn}, and StairNet \cite{woo2018stairnet} follow this principle, and significant improvement  has been reported in terms of detection accuracy.

Our work is motivated by the observation that the capacity of the current architectures for generating top-down features might not be large enough to generate strong semantics for all scales. Thus, we propose a new architecture, which deeply fuses the multiscale features for enhanced object detection. The proposed feature pyramid method, referred to as {\it semantic combining and attentive redistribution feature network} (ScarfNet), combines the multiscale feature maps using the recurrent neural networks and then redistributes the fused semantics to each level, generating the new multiscale feature maps. The structure of ScarfNet is depicted in Fig.  \ref{fig:proposed} (d). First, the bidirectional long short-term memory (biLSTM) model \cite{zhang2018attention} is used to combine the multiscale pyramidal features. Although biLSTM is widely used to extract the temporal features from the sequential data, it can effectively combine the semantics in multiscale features. Our conjecture is that compared to the convolutive fusion methods, biLSTM requires significantly reduced number of weights due to parameter sharing, and the only relevant semantic information is selectively aggregated through the gating function of biLSTM. The fused feature maps are distributed through the channel-wise attention model, generating highly semantic features tailored for each pyramid scale. The final multiscale feature maps are used for object detection. Note that our framework can be readily applied to various feature pyramid-based CNN architectures, which require strong semantic information.

In the experiments, we integrate ScarfNet with the baseline detectors including  Faster R-CNN \cite{fasterrcnn}, SSD \cite{ssd}, and RetinaNet \cite{retinanet}. The evaluation conducted over PASCAL VOC \cite{pascalvoc} and MS COCO \cite{mscoco} datasets shows that our method offers significant improvement over the baseline detectors as well as other competitive detectors in terms of detection accuracy.  Our code will be made publicly available. The contributions of our paper are summarized as follows.
\begin{itemize}
    \item We introduce a new deep architecture for closing the semantic gaps between the multiscale feature maps. The proposed ScarfNet generates new multiscale feature maps with deeply fused and redistributed semantics by using the combination of biLSTM and the channel-wise attention model.
    \item For the first time in the literature, the biLSTM is used to combine the multiscale features to incorporate strong semantics for feature pyramids. The biLSTM model can produce deeply fused semantic information using the recurrent connection over different pyramid scales. Furthermore, ScarfNet benefits from the selective information gating mechanism inherent in the biLSTM model. Due to parameter sharing, the overhead due to ScarfNet is small. In addition, ScarfNet is easy to train and is also end-to-end trainable.
\end{itemize}

\section{Related Work}
In this section, we review the basic object detectors and several existing feature pyramid methods used to decrease the semantic gap between the scales. 
\subsection{CNN-based Object Detectors}
Recently, CNN has brought an order of magnitude performance improvement in object detection. 
Various CNN-based object detectors have been proposed to date. The current object detectors can be categorized into two groups: two-stage detectors and single-stage detectors. Two-stage detectors detect the objects in two steps; finding the region proposals based on the objectness of the regions and conducting the classification and bound regression for detected region proposals. R-CNN \cite{rcnn} is the first CNN-based detector where the traditional selective search is employed to find the region proposals, and CNN is applied to the image patch in each region proposal. Fast R-CNN \cite{fastrcnn} and Faster R-CNN \cite{fasterrcnn} reduced the computation time of R-CNN by employing the region of interest (ROI) pooling for using full-image feature maps and replacing the selective search with the region proposal network (RPN). Single-stage detectors directly perform classification and box regression based on the feature maps. These detectors compute the confidence score on the object category and the regression results for the candidate boxes while sweeping the feature maps spatially. The well-known single-stage detectors include SSD \cite{ssd}, YOLO \cite{yolov1}, and YOLOv2 \cite{yolov2}. Recently, RetinaNet \cite{retinanet} has achieved the-state-of-the-art performance using ResNet \cite{resnet} as a backbone and  various latest training tricks. Refer to \cite{survey} for the comprehensive review of contemporary object detectors. 

\subsection{Object Detectors Using Multiscale Features}
Several object detectors including SSD \cite{ssd} and RetinaNet \cite{retinanet} rely on hierarchical feature pyramids to detect the objects of various sizes (Fig.  \ref{fig:comparison} (a)). 
One problem with using multiscale features directly produced by CNNs is the gap of the semantic information between them caused by the different depths of the layers passed by the input. Due to the relatively low level of abstraction for bottom-level features, detection accuracy for small objects is often limited. 
Fig. \ref{fig:comparison} (b), (c), and (d) describe the strategies proposed to overcome this problem. Fig. \ref{fig:comparison} (b) depicts the strategy of combining the multiscale features into the single high-resolution feature map with strong semantics.  HyperNet \cite{kong2016hypernet} and ION \cite{bell2016inside} improved the performance of RPN by aggregating the hierarchical features with the appropriate resizing of the feature maps. Fig. \ref{fig:comparison} (c) shows the strategy of generating highly semantic features through the top-down pathway with lateral connections. Note that the semantic information is generated through the top-down connections while the detailed spatial information is provided through lateral connections. Several detectors based on this structure include DSSD \cite{dssd}, StairNet \cite{woo2018stairnet}, TDM \cite{shrivastava2016beyond}, FPN \cite{fpn}, and RefineDet \cite{zhang2018single}. DSSD \cite{dssd} and StairNet \cite{woo2018stairnet} use the deconvolutional layer-based top-down connections for the SSD baseline \cite{ssd}. TDM \cite{shrivastava2016beyond} employs the top-down structure specified for the RPN of the Faster R-CNN \cite{fasterrcnn}. FPN \cite{fpn} uses the simplified structure using 2x up-sampling and 1x1 convolution for top-down and lateral connections, respectively. RefineDet \cite{zhang2018single} employs two-step cascade regression for top-down connections.
\begin{figure*}
	\centering
    \centerline{\includegraphics[width=0.9\textwidth]{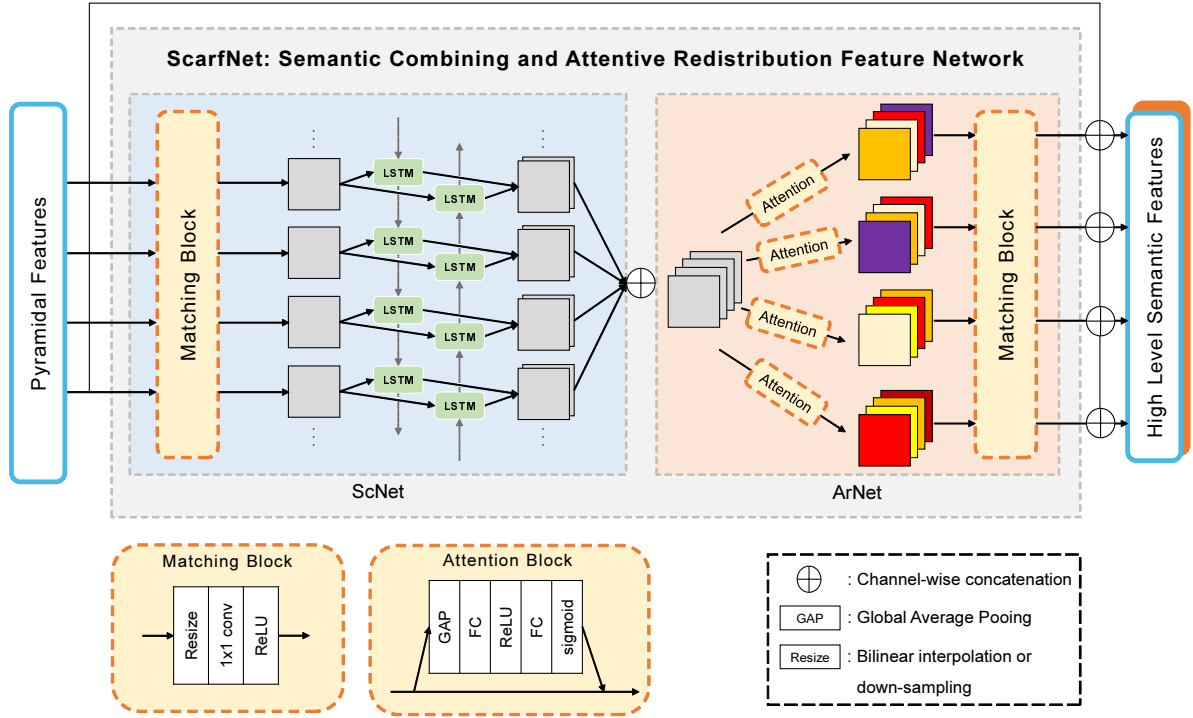}}
    \hspace{1cm}
	\caption {{\bf The overall architecture of the proposed ScarfNet:} ScarfNet consists of two modules: ScNet and ArNet. ScNet aggregates the pyramidal features obtained from the bottom-up CNN pipeline. Then, ArNet distributes the fused semantics to each pyramid level. The final high-level semantic features are generated by channel-wise concatenation between the output of ScarfNet and the original pyramidal features. The detailed structures of the matching block and the attention block are depicted in the yellow boxes.\vspace{-0.17cm}}
	\label{fig:proposed}
\end{figure*}

\section{Proposed Object Detector}
In this section, we introduce the details of the proposed ScarfNet architecture. 
\subsection{Existing Feature Pyramid Methods}
The feature pyramid-based object detectors performs detection based on the $k (> 1)$ feature maps across different pyramid levels to detect the various sizes of objects. As shown in Fig. \ref{fig:comparison} (a), the baseline detectors use the feature map $X_l$ at the $l$th pyramidal level
\begin{gather}
X_l = {B}_l\left(X_{l-1}\right)\\
\text{Detection Outputs} = {D}_l(X_{l}),
\end{gather}
where  $l=n-k+1,...,n$. Note that $X_{1:n-k}(=[X_1, X_2,...,X_{n-k}])$ are the feature maps produced by the backbone network, and $X_{n-k+1:n}$ are the bottom-up features from the subsequent convolutional layers.  ${B}_l(\cdot)$ denotes the operation performed by the $l$th convolutional layer, and ${D}_l(\cdot)$ denotes the detection subnetwork that often applies a single 3x3 convolutional layer to produce the output of classification and box regression. Due to the different depths from the input to each pyramidal feature, shallow bottom-level features suffer from the lack of semantic information. 

In order to reduce the semantic gap between different pyramid levels, several works proposed the top-down structure using lateral connections as illustrated in Fig. \ref{fig:comparison} (c). This structure propagates the high-level semantics from the top layers to the bottom layers with increased resolution while keeping the spatial high resolution through lateral connections. The $l$th feature map $X'_{l}$ generated using this method is expressed as
\begin{gather}
X'_l = {L}_l\left(X_{l}\right) \oplus {T}_l\left(X'_{l+1}\right)\\
\text{Detection Outputs} = {D}_l(X'_{l})
\end{gather}
where $l=n-k+1,...,n$. Note that $L_l( \cdot )$ is the operation for the $l$th lateral connection, and $T_l( \cdot )$ is the operation for the $l$th top-down connection. The operator $\oplus$ represents the operation of combining two feature maps, e.g.,  channel-wise concatenation and addition. Different methods (e.g., DSSD \cite{dssd}, StairNet \cite{woo2018stairnet},  TDM \cite{shrivastava2016beyond}, FPN \cite{fpn}, and RefineDet \cite{zhang2018single}) employ the slightly different structures for $L_l( \cdot )$ and $T_l( \cdot )$. While these methods promote the abstraction level for pyramidal features, they still have some limitations. 
Since the top-down connection propagates the semantic information in a unidirectional way, the semantics are not evenly distributed among all pyramid levels. As a result, the semantic gap between the pyramidal features still remains.
Next, such unilateral processing of the features has limited capacity to produce rich contextual information for increasing the semantic levels in all scales. In order to address these problems, we developed a new architecture that uses biLSTM to generate the deeply fused semantics through bi-lateral connections between all pyramid scales. In the following subsections, we present the details of our design.

\subsection{ScarfNet: Overall Architecture}
ScarfNet attempts to resolve the discrepancy of the semantic information in two steps: 1) combining the scattered semantic information using biLSTM and 2) redistributing the fused semantics back to each pyramid level using the channel-wise attention model. The overall architecture of ScarfNet is depicted in Fig. \ref{fig:proposed}. Taking the $k$ pyramidal features $X_{n-k+1:n}$ as input, ScarfNet produces the new $l$th pyramidal feature map $X'_l$ as
\begin{align}
X'_l &= \text{ScarfNet}_l\left(X_{n-k+1:n}\right) \\
&= X_l  \oplus \text{ArNet}_l(\text{ScNet}(X_{n-k+1:n})) 
\label{eq:arsc}
\end{align}
\vspace{-0.9cm}
\begin{align}
\text{Detection Outputs} = {D}_l(X'_{l})\quad
\end{align}
where $l=n-k+1,...,n$.
As in (\ref{eq:arsc}), ScarfNet consists of two subnetworks: semantic combining network (ScNet) and attentive redistribution network (ArNet).
First, ScNet merges the pyramidal features $X_{n-k+1:n}$ through biLSTM and produces the output features with the fused semantics. 
Second, ArNet collects the output features from biSLTM and applies the channel-wise attention model to produce highly semantic multiscale features, which are concatenated to the original pyramidal features. 
Finally, the resulting feature maps are individually processed by the detection subnetwork ${D}_l(\cdot)$ to produce the results for object detection.
\subsection{Semantic Combining Network (ScNet)}
\begin{figure}
	\centering
    \centerline{\includegraphics[width=0.48\textwidth]{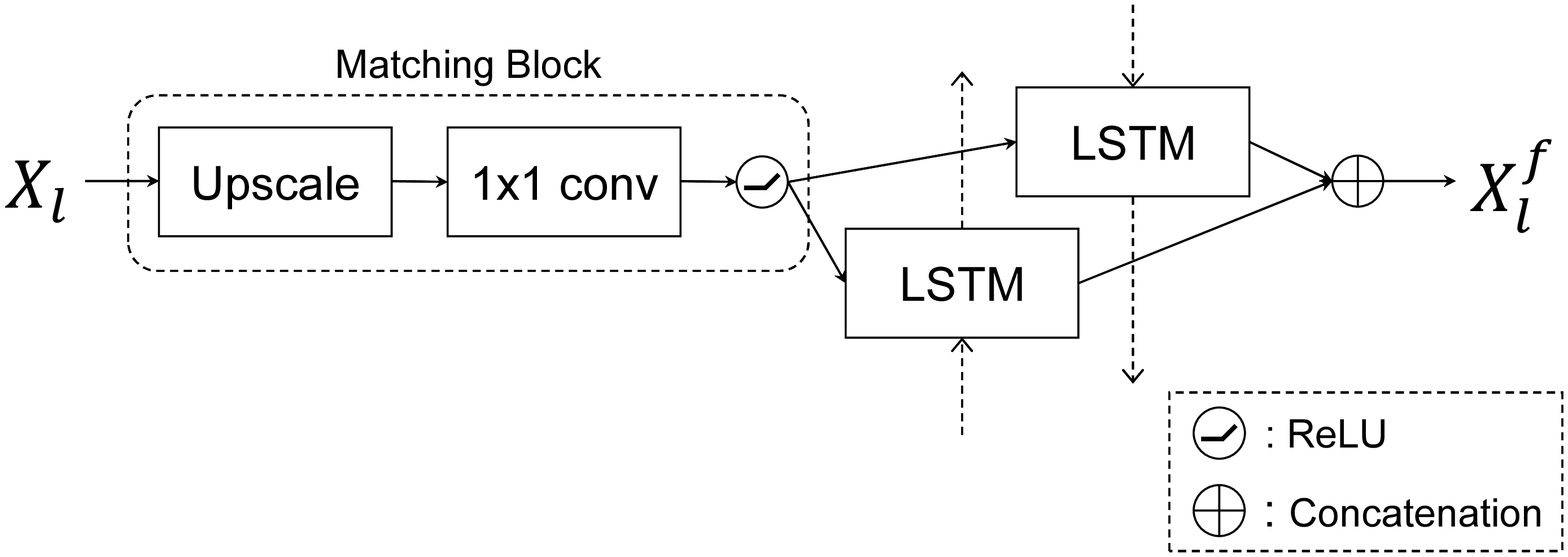}}
	\caption {{\bf The structure of ScNet}: The matching block and biLSTM are applied to generate the fused feature map $X_l^f$. Note that the matching block applies bilinear interpolation and 1x1 convolution to make the spatial and channel dimensions equal for the inputs to biLSTM. }
	\label{fig:ScNet}
\end{figure}

 The feature maps $X^f_{n-k+1:n}$ produced by ScNet are obtained as follows:
\begin{gather}
X^f_{n-k+1:n}  = \text{ScNet}(X_{n-k+1:n}),
\end{gather}
where $X^f_{l}$ is the output feature map for the $l$th layer. Fig. \ref{fig:ScNet} depicts the detailed structure of ScNet.
ScNet uniformly fuses the semantics scattered in different pyramid levels  using biLSTM. The biLSTM model can selectively fuse the contextual information in multiscale features through the gating function.
As shown in Fig.~\ref{fig:ScNet}, ScNet consists of the matching block and the biLSTM block. The matching block first resizes the pyramidal features $X_{n-k+1:n}$ such that they have the same size as the largest pyramidal feature. Then, it adjusts the channel dimension of the input using the 1x1 convolutional layer. As a result, the matching block produces the feature maps of the same spatial and channel dimensions for biLSTM. Note that the resizing operation is performed by bilinear interpolation. The biLSTM model used in SCNet follows the structure of \cite{xingjian2015convolutional}, which has significantly saved the computation time by using convolutional layers for input connection and computing the gating parameters based on the results of global average pooling. The operations performed by biLSTM in \cite{xingjian2015convolutional} are summarized as follows:
\begin{gather}
\bar{X}_l = GlobalAveragePooling(X_l)\\
\bar{X}^f_{l-1} = GlobalAveragePooling(X^f_{l-1})\\
i_l = \sigma\left(W_{xi}\bar{X}_l +W_{x^f i}\bar{X}^f_{l-1} +b_i \right)\\
f_l = \sigma\left(W_{xf}\bar{X}_l +W_{x^f f}\bar{X}^f_{l-1} +b_f \right)\\
o_l = \sigma\left(W_{xo}\bar{X}_l +W_{x^f o}\bar{X}^f_{l-1} +b_o \right)\\
G_l = \tanh\left(W_{xc}*X_l +W_{x^f c}*X^f_{l-1} +b_c \right)\\
C_t = X_l\circ C_{l-1} + i_l \circ G_l\\
X^f_l = o_l \circ \tanh\left(C_l\right),
\end{gather}
where $\circ$ denotes the Hadamard product. The state update of biLSTM is conducted in both forward and backward directions. Note that we only provide the forward update, and the equations are similar for the backward update.

\begin{figure} [tbh]
	\centering
    \centerline{\includegraphics[width=0.48\textwidth]{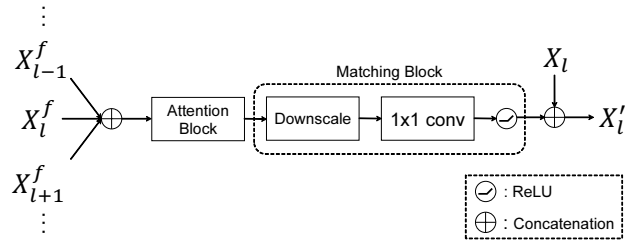}}
	\caption {{\bf The structure of ArNet}: ArNet concatenates the fused feature maps $X_{n-k+1:n}^f$ and applies the channel-wise attention. Then, the spatial and channel dimensions of the resulting feature maps are adjusted by the matching block.   }
	\label{fig:ArNet}
\end{figure}
\renewcommand{\arraystretch}{1.1}
\begin{table*}[ht]
\begin{center}
\begin{adjustbox}{width=0.6\textwidth}
\begin{tabular}{c|c|c|c|c}
\Xhline{4\arrayrulewidth}
\multirow{2}*{Method} &\multirow{2}*{Backbone} & \multirow{2}*{Input size} & \multicolumn{2}{c}{mAP (\%)}\\ \cline{4-5}
&&&VOC 2007& VOC 2012  \\
\hline\hline
StairNet \cite{woo2018stairnet} & VGG-16                   & $300\times300$ &78.8&76.4 \\
Faster R-CNN \cite{fasterrcnn} & VGG-16     & $\sim 1000\times 600$ & 73.2 & 70.4 \\
ION \cite{bell2016inside}          & VGG-16     & $\sim 1000\times 600$ & 76.5 & 76.4 \\
\hline
SSD300* \cite{ssd}  & VGG-16                   & $300\times300$ &77.5& 75.8\\
\textbf{Scarf SSD300 (ours)} & VGG-16& $300\times300$ & 79.4 & 77.2\\ 
\hline
SSD512* \cite{ssd}   & VGG-16                   & $512\times512$ &79.8& 78.5\\
\textbf{Scarf SSD512 (ours)} & VGG-16& $512\times512$ & {\bf 81.6} &  {\bf 79.8} \\ 
\hline\hline
SSD321 \cite{dssd}   & ResNet-101               & $321\times321$ &77.1& 75.4\\
SSD513 \cite{dssd}   & ResNet-101               & $513\times513$ &80.6& 79.4\\
DSSD321 \cite{dssd}  & ResNet-101               & $321\times321$ &78.6& 76.3\\
DSSD513 \cite{dssd}  & ResNet-101               & $513\times513$ &81.5& 80.0\\
R-FCN \cite{rfcn}    & ResNet-101 & $\sim 1000\times 600$ & 80.5 & 77.6 \\
\hline
$\text{Faster R-CNN}^{\dagger}$ \cite{fasterrcnn}  & ResNet-101               & $\sim833\times500$ &81.1& - \\
\textbf{Scarf Faster R-CNN (ours)} &ResNet-101& $\sim833\times500$&82.3& - \\ 
\hline
$\text{RetinaNet500}^{\dagger}$ \cite{retinanet}  & ResNet-101               & $\sim833\times500$ &83.0& - \\
\textbf{Scarf RetinaNet500 (ours)} &ResNet-101& $\sim833\times500$&  \textbf{83.5} & - \\ 
\Xhline{4\arrayrulewidth}
\end{tabular}
\end{adjustbox}
\end{center}
\caption{\textbf{PASCAL VOC 07/12 detection results:} The detection results for VOC 2007 are evaluated on {\it VOC 2007 test set} after trained on {\it VOC 2007 trainval} and {\it VOC 2012 trainval}. Those for VOC 2012 are evaluated on {\it VOC 2012 test set} when trained on {\it VOC 2007 test}, {\it VOC2007 trainval}, and {\it VOC 2012 trainval} sets.}
\label{table:voc}
\end{table*}
\renewcommand{\arraystretch}{1}

\subsection{Attentive Redistribution Network (ArNet)}
The ArNet aims to produce the high-level semantic feature map, which is concatenated with the original pyramidal feature map $X_l$ as follows:
\begin{gather}
X'_l  = X_l  \oplus \text{ArNet}_l(X^f_{n-k+1:n}),
\end{gather}
where operator $\oplus$ denotes the channel-wise concatenation. The detailed structure of ArNet is depicted in Fig. \ref{fig:ArNet}. ArNet  concatenates the outputs $X^f_{n-k+1:n}$ of biLSTM and apply the channel-wise attention to them. The attention weights are obtained by constructing the 1x1 vector using global average pooling \cite{senet} and passing it through two fully connected layers followed by the sigmoid function. Note that this channel-wise attention model allows for selective propagation of the semantics to each pyramid level. Once the attention weights are applied, the matching block down-samples the resulting feature maps to the original size of the pyramidal features and applies 1x1 convolution to match the channel dimensions with those of the original pyramidal features. Finally, the output of the matching block is concatenated with the original feature $X_l$ to produce the highly semantic feature $X'_l$.  
\section{Experiments}
We evaluated the performance of the proposed ScarfNet model by comparing our detector with other multiscale detection methods and conducting an extensive performance analysis to understand the behavior of our architecture. 
\subsection{Experimental Setup}
ScarfNet was applied to the baseline object detectors including Faster R-CNN \cite{fasterrcnn}, SSD \cite{ssd}, and RetinaNet \cite{retinanet}. In the case of Faster R-CNN and RetinaNet, we replaced the original FPN part with the feature generation by ScarfNet. 
We compared our method with the baselines, Faster R-CNN \cite{fasterrcnn}, SSD \cite{ssd}, and RetinaNet \cite{retinanet}, as well as the other competitive algorithms including ION \cite{bell2016inside}, R-FCN \cite{rfcn}, DSSD \cite{dssd}, and StairNet \cite{woo2018stairnet}. We measured the mean average precision (mAP) in \% on three widely used datasets for object detection benchmark: PASCAL VOC 2007, PASCAL VOC 2012 \cite{pascalvoc}, and MS COCO \cite{mscoco}.
\begin{table*}[t]
\centering
\begin{adjustbox}{width=0.9\textwidth}
\begin{tabular}{c|c|c|c|c|c|ccc|ccc}
\Xhline{4\arrayrulewidth}\xrowht{5pt}
Method & Network  & Backbone & Module & Input size &fps& $\text{AP}$ & $\text{AP}_{50}$ & $\text{AP}_{75}$& $\text{AP}_S$& $\text{AP}_M$& $\text{AP}_L$\\
\hline\hline\xrowht{5pt}

\multirow{4}{*}{\textit{two-stage}}&\multirow{2}{*}{Faster R-CNN* \cite{fasterrcnn}}&  ResNeXt-101    & FPN & $\sim 833\times500$ &15.3&37.6&59.1&40.7&19.2&41.8&52.3\\
&&  ResNeXt-101  & FPN &$\sim 1333\times800$&10.3&41.9& 63.9& 45.9&25.0&45.3&52.3\\
\cline{2-12}\xrowht{5pt}
&\multirow{2}{*}{\textbf{Scarf Faster R-CNN (ours)}}  & ResNeXt-101 &SCARF&$\sim 833\times500$  &13.8&38.5&59.9&41.5&19.1&42.9&\textbf{54.1}\\
&   &ResNeXt-101&SCARF& $\sim 1333\times800$  &8.9 &\textbf{42.8}&\textbf{64.3}&\textbf{47.1}&\textbf{26.0}&\textbf{45.7}&52.9\\
\hline\hline\xrowht{5pt}
\multirow{8}{*}{\textit{one-stage}}&SSD513 \cite{dssd} &  ResNet-101  & - &$513\times513$&12.5&31.2&50.4&33.3&10.2&34.5&49.8\\
&DSSD513 \cite{dssd} &  ResNet-101  & DSSD &$513\times513$&10.0&33.2&53.3&35.2&13.0&35.4&51.1\\
\cline{2-12}\xrowht{15pt}
&\textbf{Scarf SSD513 (ours)}& ResNet-101 &SCARF&$513\times513$&11.5&\textbf{34.5}&\textbf{54.1}&\textbf{36.3}&\textbf{15.1}&\textbf{36.1}&\textbf{51.6}\\
\cline{2-12}\xrowht{5pt}
&\multirow{2}{*}{RetinaNet \cite{retinanet}}&  ResNet-101    & FPN & $\sim 833\times500$ &15.4&34.4&53.1&36.8&14.7&38.5&49.1\\
&&  ResNeXt-101  & FPN &$\sim 1333\times800$&9.3&40.8&61.1&44.1&24.1&44.2&51.2\\
\cline{2-12}\xrowht{5pt}
&\multirow{2}{*}{\textbf{Scarf RetinaNet (ours)}}  & ResNet-101 &SCARF&$\sim 833\times500$  &13.6&35.1&53.8&37.7&15.8&38.7&49.0\\\xrowht{5pt}
 & &ResNeXt-101&SCARF& $\sim 1333\times800$  &8.4&\textbf{41.6}&\textbf{62.0} &\textbf{44.6}&\textbf{24.5}&\textbf{45.5}&\textbf{52.3}\\
\Xhline{4\arrayrulewidth}
\end{tabular}
\end{adjustbox}
\caption{\textbf{Detection results on \textit{MS COCO test-dev} dataset}: The symbol ``*" indicates our re-implemented results. The expression ``$\sim x\times y$" means re-scaling of the input image introduced in the original RetinaNet paper.}
\label{table:coco}
\end{table*}

\subsection{Network Configuration}
The advantage of ScarfNet is that there are not many hyperparameters to be determined. Note that the spatial dimensions of the feature maps are readily determined based on those of the baseline detectors. The channel dimensions of the intermediate feature maps are fixed over the pipeline between two matching blocks in ScNet and ArNet. Thus, we only need to determine this channel dimension. According to our empirical results, we set the channel dimension to 256.  

\subsection{Performance Evaluation}
\subsubsection{PASCAL VOC Results}
\hspace*{5mm}{\bf Training on PASCAL VOC 2007 Dataset}:  The object detectors under consideration were trained with the {\it VOC 2007 trainval} and the {\it VOC 2012 trainval} sets and evaluated with the {\it VOC 2007 test} set. When ScarfNet was combined with the SSD baseline, we trained our model over 120k iterations ($\sim$240 epochs). We used a learning rate of $10^{-3}$ for the first $80$k iterations, $10^{-4}$ for the next $20$k iterations, and $ 10^{-5}$ for the last $20$k iterations. The mini-batch size was set to $32$, the momentum for the stochastic gradient descent (SGD) update was set to $0.9$, and the weight decay was set to $0.0005$.
When our method was combined with the RetinaNet baseline, we used a learning rate of $5\times10^{-3}$ for the first $60$k iterations, $5\times10^{-4}$ for the next $20$k iterations, and  $5\times10^{-5}$ for the last $10$k iterations. Other parameters were equally set except for the weight decay of $0.0001$. \\
\hspace*{5mm}{\bf Training on PASCAL VOC 2012 Dataset}: The object detectors were trained with the {\it VOC 2007 trainval}, the {\it VOC 2007 test} and the {\it VOC 2012 trainval} sets and evaluated with the {\it VOC 2012 test} set. When our model was applied to the SSD baseline, a total of 200k iterations were run with the same training parameters as in the VOC 2007 case. Note that we used a learning rate of $10^{-3}$ for the first $120$k iterations, $10^{-4}$ for the next $40$k iterations, and $ 10^{-5}$ for the rest. 
\\
\hspace*{5mm}{\bf Performance Comparison}: Table \ref{table:voc} shows the mAP performance of the object detectors under comparison evaluated on the PASCAL VOC 2007 and 2012 test sets. For both PASCAL 2007 and 2012 cases, the semantic features generated by ScarfNet offer a significant performance gain over the baseline detectors. In the case of PASCAL VOC 2007, the proposed method achieves 1.9\%, 1.8\%, and 1.2\% mAP gains over the SSD300, SSD512, and Faster R-CNN baselines, respectively. The proposed method also outperforms the RetinaNet baseline by 0.5\%. Since the RetinaNet baseline employs the top-down structure based on FPN \cite{fpn}, we conclude that the features generated by our method are superior to those generated by FPN. Our object detector also achieves better performance than the other competing algorithms including, StairNet \cite{woo2018stairnet}, DSSD \cite{dssd}, ION \cite{bell2016inside}, R-FCN \cite{rfcn}. 
Although the detection accuracy with the PASCAL VOC 2012 dataset is slightly degraded compared to that with PASCAL VOC 2017, the tendency of the detection results observed for PASCAL VOC 2007 remains the same. 
Note that the proposed detector maintains the performance gain of 1.4\% and 1.3\% mAP over the SSD300 and SSD500 baselines, respectively.


\begin{table}[t]
\centering
\begin{adjustbox}{width=0.48\textwidth}
\begin{tabular}{c|c|c}
\Xhline{4\arrayrulewidth}
 &Method&mAP  \\
 \hline \hline\xrowht{7pt}
 \multirow{3}{*}{\begin{tabular}[c]{@{}c@{}}Ablation \\ study\end{tabular}}&Basedline (SSD) & 77.5 \\
 \cline{2-3}\xrowht{7pt}
 &biLSTM & 79.1 \\
 \cline{2-3}\xrowht{7pt}
 &biLSTM + channel-wise attention&\textbf{79.4}\\
\hline\hline\xrowht{7pt}
 \multirow{4}{*}{\begin{tabular}[c]{@{}c@{}}Other fusion strategy \\ (used with channel-wise attention)\end{tabular}} &1x1 conv.-based fusion&78.9  \\
 \cline{2-3}\xrowht{7pt}
 &uniLSTM & 78.7 \\
 \cline{2-3}\xrowht{18pt}
 &\begin{tabular}[c]{@{}c@{}}Top-down structure\\with lateral connections\end{tabular}& 78.6 \\
 
\Xhline{4\arrayrulewidth}
\end{tabular}
\end{adjustbox}
\caption{\textbf{Results of ablation study on \textit{VOC 2007 test} dataset.}}
\label{table:ablation}
\end{table}

\renewcommand{\arraystretch}{1.3}
\begin{table}
\begin{center}
\begin{adjustbox}{width=0.42\textwidth}
\begin{tabular}{c|c||c|c}
\Xhline{4\arrayrulewidth}
\multicolumn{2}{l||}{} & \multicolumn{2}{c}{Semantic feature generation strategy}   \\ \cline{3-4}
\multicolumn{2}{l||}{} & \qquad Addition \quad\quad & Concat. \\ 
 \hline \hline
\multicolumn{1}{c|}{\multirow{5}{*}{ \pbox{20cm}{Channel \\ dimension}} }    & 64  &    78.3 & 78.8   \\ \cline{2-4}
\multicolumn{1}{c|}{}    & 128  & 78.6 & 79.1 \\ \cline{2-4}
\multicolumn{1}{c|}{}    & 256  & 79.1 & \bf{79.4} \\\cline{2-4}
\multicolumn{1}{c|}{}    & 512  & 79.5 & 79.2 \\\cline{2-4}
\multicolumn{1}{c|}{}    & 1024 & 79.4 & 79.2 \\
\Xhline{4\arrayrulewidth}
\end{tabular}

\end{adjustbox}
\end{center}
\caption{mAP (\%) performance for various combinations of channel dimension and semantic feature generation strategy when evaluated on {\it VOC 2007 test set}}
\label{table:ablation2}
\end{table}
\renewcommand{\arraystretch}{1}

\subsubsection{COCO Results}
\hspace*{5mm}{\bf Training}: The object detectors under comparison were trained with the {\it MS COCO trainval35k} split \cite{bell2016inside} (union of 80k images from the training set and a random 35k subset of images from 40k image val split). The evaluation was performed using the {\it MS COCO test-dev}. To train the proposed structure based on RetinaNet \cite{retinanet}, we used the learning rate of $10^{-2}$ for the first $60$k iterations, $10^{-3}$ for the next $20$k iterations, and $ 10^{-5}$ for the last $20$k iterations. The mini-batch size was set to $16$, the momentum was set to $0.9$, and the weight decay was set to $0.0001$.\\
\hspace*{5mm}{\bf Performance comparison}: Table \ref{table:coco} provides the detection accuracy of the algorithms tested on the MS COCO dataset. The experiment was conducted on various baseline detectors and feature pyramid modules. The proposed Scarf SSD513 and Scarf RetinaNet achieve the significant performance gain over the baselines. Our method beats  the Faster R-CNN  baseline \cite{fasterrcnn} by 0.9\% AP. Note also that the Scarf SSD513 achieves 1.3\% performance gain over DSSD513 and  the Scarf RetinaNet offers the 0.8\% performance gain over the RetinaNet baseline  \cite{retinanet}.

\begin{figure*}[t]
\begin{center}
\includegraphics[width=0.55\textwidth]{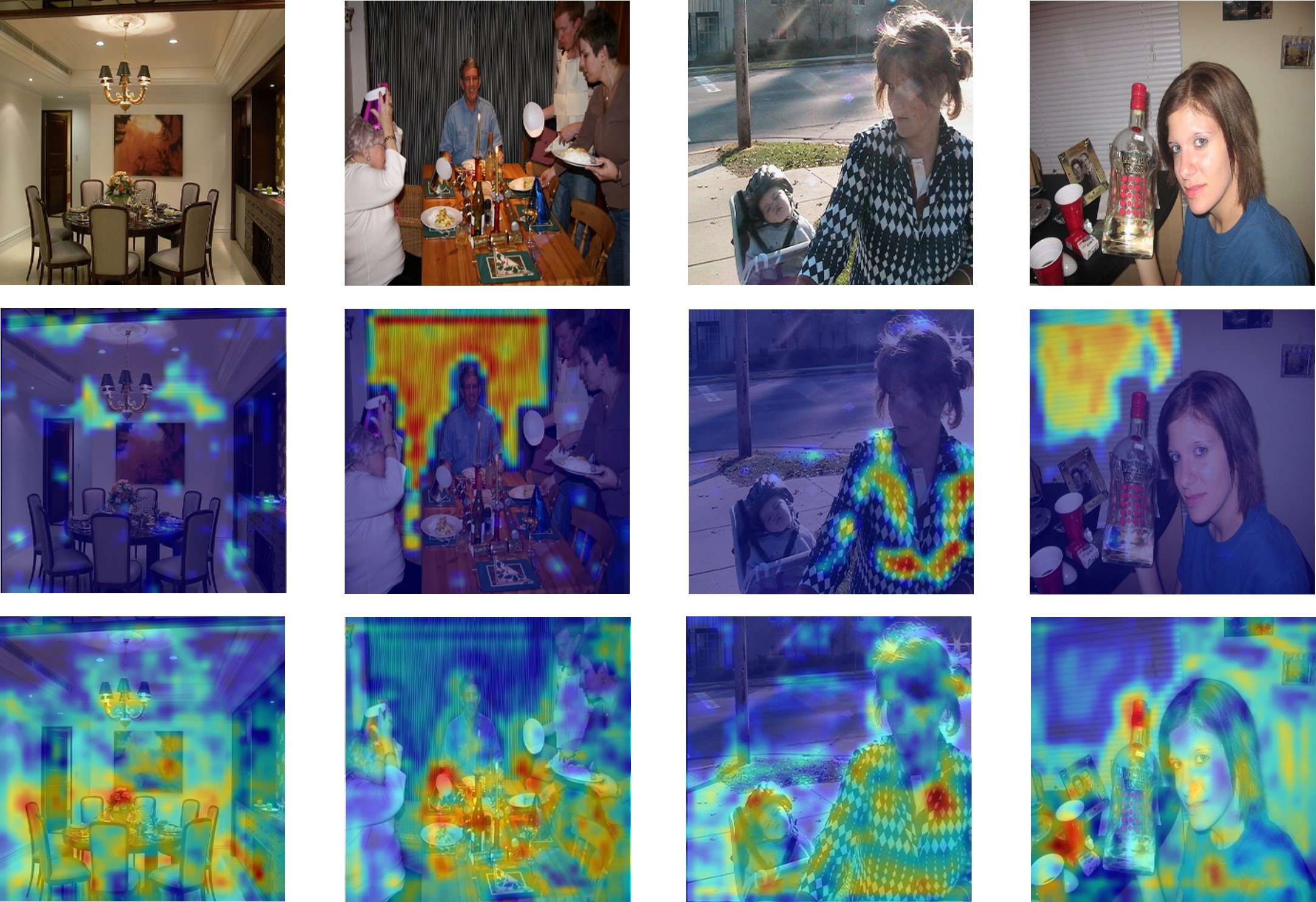}
\end{center}
\caption{\textbf{Visualization of the feature map}: (top row) input image, (middle row) {\tt conv4\_3} layer feature ($X_1$) from the feature pyramid in SSD300, (bottom row) {\tt conv4\_3} layer feature ($X'_1$) generated from ScarfNet. Since the {\tt conv4\_3} layer feature map $X_1$ is shallow, it fails to place strong activation properly on the objects. On the contrary,  the semantic feature generated by ScarfNet captures the characteristics of the objects well. }
\label{fig:featvis}
\end{figure*}

\subsection{Performance Analysis}
\subsubsection{Ablation Study}

{\bf Benefits of biLSTM}: It is worth investigating the effectiveness of biLSTM and channel-wise attention for fusing the multi-scale features. Table \ref{table:ablation} shows how the performance of our method improves as we add bi-LSTM and channel-wise attention to the baseline one by one. We see that the biLSTM offers the 1.6\% AP gain over the baseline and combination of biLSTM and channel-wise attention adds 1.9\% AP gain. 
Table \ref{table:ablation} also compares the different fusion strategies including the 1x1 convolutional layer, the top-down structure, and the unidirectional LSTM. Our biLSTM achieves better performance than the others; thus, parameter sharing, gating units, and bilateral processing of biLSTM effectively control high-level information to reduce the subtle semantic gap between the hierarchical features.\\

{\bf Network Parameter Search} We need to determine the channel dimension of the intermediate feature maps. We should also determine whether the element-wise addition or channel-wise concatenation is better in combining the output of ScarfNet with the original feature pyramid. Table \ref{table:ablation2} shows the evaluation of the performance of our detector for various combinations of channel dimensions (64, 128, 256, 512 versus 1024) and feature combining strategies (element-wise addition versus channel-wise concatenation). According to Table \ref{table:ablation2}, the combination of 512 channel dimensions with element-wise addition leads to the best detection accuracy. However, using 512 channels significantly increases the computational complexity of the entire network; thus, we chose 256 channel dimensions with channel-wise concatenation.

\subsubsection{Feature Visualization}
We investigated the effectiveness of ScarfNet via feature visualization. Fig. \ref{fig:featvis} compares the original pyramidal feature map $X_1$ of the largest size (middle row) with the semantic feature map $X'_1$ from ScarfNet (bottom row). In order to obtain the heat map, we took the channel with the highest average activation in the spatial domain. Due to the lack of the semantic cues in the original feature map $X_1$, it often fails to activate on the objects properly. On the contrary, the feature map $X'_1$ has strong activation on the entire region occupied by the objects, which would lead to an improvement in the overall detection performance. 

\section{Conclusions}
In this study, we developed a deep architecture that generates multiscale features with strong semantics to reliably detect the objects in various sizes. Our ScarfNet method transforms the pyramidal features produced by the baseline detector into evenly abstract features. ScarfNet fuses the pyramidal features using biLSTM and distributes the semantics back to each multiscale feature. We verified through experiments conducted with PASCAL VOC and MS COCO datasets that the proposed ScarfNet method significantly increases the detection performance over the baseline detectors. Our object detector achieves the state-of-the-art performance on the PASCAL VOC and COCO benchmarks.

{\small
\bibliographystyle{ieee_fullname}
\bibliography{egbib}
}

\end{document}